\documentclass[11pt]{article}

\usepackage[preprint]{acl}

\usepackage{times}
\usepackage{latexsym}

\usepackage[T1]{fontenc}
\usepackage[utf8]{inputenc}
\usepackage{booktabs} 
\usepackage{amsmath} 
\usepackage{multirow}
\usepackage{amssymb}
\usepackage{microtype}

\usepackage{inconsolata}

\usepackage{graphicx}
\graphicspath{{../}}

\title{UniMem: Complementary Episodic-to-Parametric Memory for Boundary-Agnostic Task Streams}
\author{
  Siyu Xia$^{1,2}$\thanks{~Equal contribution.}, 
  Chenheng Zhang$^{3}\footnotemark[1]$,
  Yanting Wu$^{1,2}$,
  Haoxuan Li$^{4}$,
  Jiajun Chai$^{5}$,
  Xiaohan Wang$^{5}$, \\
  \textbf{Guojun Yin$^{5}$},
  \textbf{Wei Lin$^{5}$},
  \textbf{Zhouchen Lin$^{3}$},
  \textbf{Haifeng Zhang$^{1,2}$\thanks{~Corresponding authors: \texttt{haifeng.zhang@ia.ac.cn}, \texttt{jun.wang@cs.ucl.ac.uk}}}, 
  \textbf{Jun Wang$^{6}\footnotemark[2]$}
}
\begin{document}
\maketitle
\begingroup
  \renewcommand{\thefootnote}{} % 清空系统的脚注标号
  \footnotetext{%
    \hspace{-0.8em}% 抵消默认缩进
    $^{1}$Institute of Automation, Chinese Academy of Sciences, Beijing, China \hspace{6pt}
    $^{2}$School of Artificial Intelligence, University of Chinese Academy of Sciences, China \hspace{6pt}
    $^{3}$State Key Lab of General Artificial Intelligence, School
of Intelligence Science and Technology, Peking University \hspace{6pt}
    $^{4}$ Institute for Artificial Intelligence, Peking University \hspace{6pt}
    $^{5}$Meituan \hspace{6pt}
    $^{6}$AI Centre, Department of Computer Science, University College London, London, UK%
  }%
\endgroup
\begin{abstract}
Memory is essential for LLM agents to accumulate task experience and reuse task-specific execution strategies
. However, real-world deployment over boundary-agnostic and evolving task streams exposes a fundamental stability-plasticity dilemma. External retrieval-based memory can rapidly absorb new evidence, but it often fails to internalize recurring execution patterns and incurs inference-time retrieval overhead. Parametric memory enables stable and efficient execution once learned, but typically relies on explicit task boundaries and fixed parameter budgets. Inspired by the human brain, which balances plasticity and stability through complementary episodic storage and gradual consolidation, we propose \textbf{UniMem}, a self-routing framework for autonomous memory management. UniMem uses learnable routing tokens as memory controllers, enabling adaptive coordination between complementary memory pathways: novel or sparse tasks are retained in an episodic buffer for retrieval-augmented execution, while recurring and reliable patterns are consolidated into expandable parametric memory. By decoupling task identification from task execution with routing tokens and parametric memory blocks, UniMem expands memory on demand without task labels during deployment or uncontrolled parameter growth. Experiments on long-horizon streaming task sequences show that UniMem consistently outperforms baselines while maintaining execution fidelity, achieving an average gain of 4.0 EM points across three backbone models.
\end{abstract}

\section{Introduction}

Memory is a core capability for LLM agents, enabling them to accumulate task experience and reuse task-specific execution strategies across interactions~\citep{hu2026memoryageaiagents,zhang2026rethinking,zhao2024expelllmagentsexperiential,wu2026tokmemonetokenproceduralmemory}. Existing approaches to building memory broadly fall into two categories: external memory and parametric memory~\citep{wu2025humanmemoryaimemory}. External memory, represented by Retrieval-Augmented Generation (RAG)~\citep{oche2025systematicreviewkeyretrievalaugmented}, stores instructions, demonstrations, or experiences outside model parameters and retrieves them during inference. Its main advantage is high plasticity: new evidence can be incorporated immediately without gradient updates, making it suitable for diverse and mixed tasks. In contrast, parametric memory internalizes reusable task knowledge into continuous trainable spaces. A widely adopted family of approaches is Parameter-Efficient Fine-Tuning (PEFT), including LoRA- and prompt-based variants~\citep{wang2023orthogonalsubspacelearninglanguage,li2024mixloraenhancinglargelanguage}, which store task knowledge in lightweight modules while keeping the backbone frozen. Such parametric memory provides stable and efficient execution once the corresponding task pattern has been sufficiently learned.

\begin{figure*}[htbp]
\centering
\includegraphics[width=\textwidth]{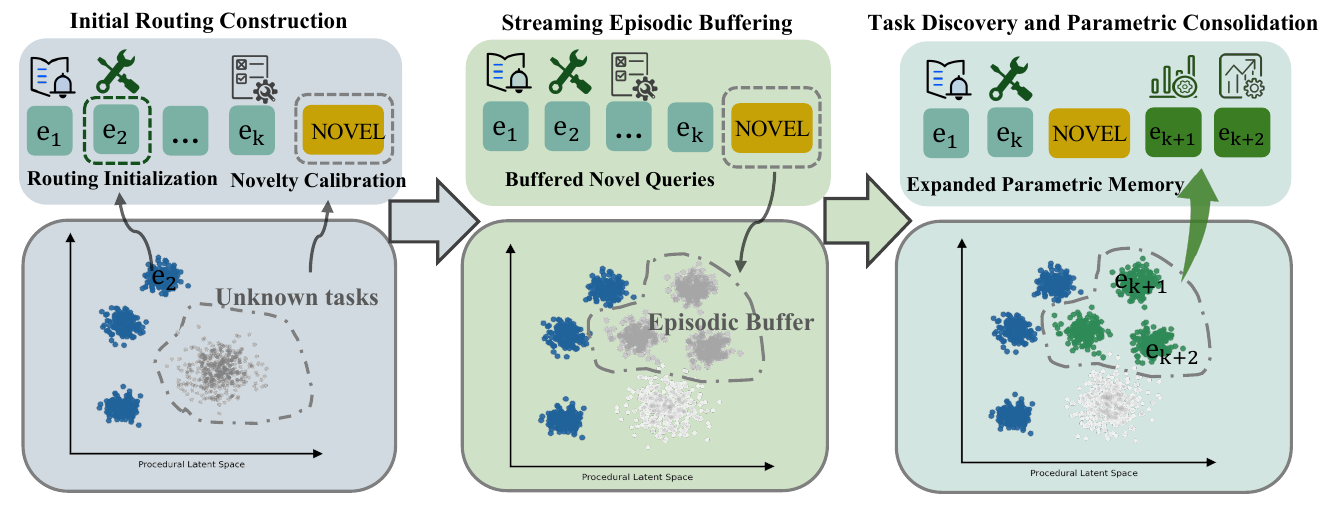}
\caption{
The episodic-to-parametric conceptual workflow of \textbf{UniMem}. 
\textbf{Initial Routing Construction} initializes known-task routing embeddings and calibrates the novelty sentinel $\mathbf{e}_{\texttt{NOVEL}}$. 
\textbf{Episodic Buffering} stores queries routed to $\mathbf{e}_{\texttt{NOVEL}}$ for retrieval-augmented execution. 
\textbf{Parametric Consolidation} clusters recurring buffered patterns and converts them into new routing embeddings with associated parametric memory units.
}
\label{fig:overall_method}
\end{figure*}

However, practical deployment exposes a harder setting: LLM agents must operate over streaming tasks that are heterogeneous, boundary-agnostic, and continuously evolving~\citep{Delange_2021,yang2025continual,bidaki2025onlinecontinuallearningsystematic}. In this setting, relying on either external or parametric memory alone leads to a stability-plasticity dilemma. A purely external-memory solution remains flexible for mixed inputs, but it often shows limited generalization and suboptimal performance when task patterns recur, since retrieved examples or instructions do not reliably internalize reusable execution behavior into the model; it also incurs retrieval and context-processing overhead during inference. A purely parametric solution offers stable internalization, but it typically requires explicit task boundaries and a fixed parameter budget, making it difficult to accommodate heterogeneous streams whose task composition and scale change over time.

Complementary Learning Systems (CLS) theory provides a natural perspective for resolving this tension \citep{McClelland1995WhyTA,kumaran2016learning}. CLS argues that a single memory system cannot simultaneously achieve high plasticity for rapidly absorbing new experiences and high stability for preserving reusable knowledge without interference. Instead, the human brain addresses this stability-plasticity trade-off through complementary memory systems: the hippocampus supports fast episodic storage of new experiences, while the neocortex gradually consolidates recurring structures into stable long-term knowledge. This human-memory mechanism motivates our problem formulation: how can an LLM agent autonomously coordinate episodic retrieval with selective parametric consolidation in an boundary-agnostic task stream?

To address this challenge, we propose \textbf{UniMem}, a unified memory framework for open-world autonomous memory expansion. As shown in Figure~\ref{fig:overall_method}, UniMem operationalizes the complementary-memory perspective through a self-routing episodic-to-parametric lifecycle. Its central mechanism is a set of learnable routing tokens that serve as memory controllers, learning task-discriminative signals to select an appropriate memory pathway for each query. Given an incoming query, UniMem compares it with consolidated task tokens and a calibrated novelty sentinel $\mathbf{e}_{\texttt{NOVEL}}$; this routing decision directly determines the subsequent memory access. Queries matched to known tokens activate the corresponding parametric memory block $\mathbf{P}_k$ for stable execution, whereas novel or uncertain queries are sent to an episodic buffer for retrieval-augmented support, avoiding premature parameter allocation. As evidence accumulates in the buffer, recurring patterns are clustered without task labels, filtered by a quality gate, and promoted into new memory units by allocating a fresh routing token paired with a parametric memory block. In this way, UniMem learns an adaptive memory management policy, retaining sparse evidence externally while autonomously consolidating reliable recurring patterns into expandable parametric memory.

In summary, our primary contributions are as follows:
\begin{itemize}
    \item \textbf{Heterogeneous Streaming Tasks as a Memory Challenge:} We formulate heterogeneous, boundary-agnostic, and evolving task streams as a practical memory-management challenge for LLM-agent deployment, where relying solely on external or parametric memory exposes a stability-plasticity dilemma.
    
    \item \textbf{Human-Brain-Inspired Unified Memory Framework:} Inspired by Complementary Learning Systems theory, we propose UniMem, an automatic memory system that routes rare tasks to episodic memory and consolidates frequent patterns into parametric memory.
    
    \item \textbf{Extensive Empirical Validation:} Evaluations across diverse long-horizon streaming settings show that UniMem consistently outperforms existing baselines while maintaining robust task execution and memory stability during autonomous memory expansion.
\end{itemize}

\section{Related Work}

\subsection{Memory-Augmented LLMs}
Current memory augmentation paradigms broadly fall into two categories: external memory and parameter-based internal memory. External memory stores task-relevant knowledge outside model parameters and injects it through the context window as a dynamic workspace, with recent research exploring strategic knowledge extraction from agent experiences~\citep{shinn2023reflexionlanguageagentsverbal,zhao2024expelllmagentsexperiential}. To handle expanding external contexts over long horizons, adaptive management mechanisms like A-MEM~\citep{xu2025amemagenticmemoryllm} and MemAgent~\citep{yu2025memagentreshapinglongcontextllm} have been introduced, while structural approaches organize historical contexts into logic-aware graphs for complex reasoning~\citep{rasmussen2025zeptemporalknowledgegraph, zhang2025gmemorytracinghierarchicalmemory}. Although effective within their respective memory structures, these methods are typically structure-specific and lack an explicit mechanism for transforming episodic evidence into consolidated parametric memory.

Parameter-based adaptation methods reduce context overhead by storing task-specific behavior in lightweight trainable components while keeping most backbone parameters frozen. Prefix-Tuning~\citep{li2021prefixtuningoptimizingcontinuousprompts} optimizes continuous prefix vectors, LoRA and QLoRA~\citep{hu2021loralowrankadaptationlarge,dettmers2023qloraefficientfinetuningquantized} inject low-rank updates, and steering methods use small latent transformations to control model behavior~\citep{han2024word,zhanglanguage}. Related work further explores instance-dependent prompt generation~\citep{wu2022idpginstancedependentpromptgeneration}, ultra-compact prompt anchors~\citep{liu2025needonecapsuleprompt}, self-updatable memory pools~\citep{wang2024memoryllmselfupdatablelargelanguage}, and reusable KV caches for reasoning~\citep{gupta2026reasoncacheteachingllmsreason}. However, these methods usually assume a predefined task set or a fixed adaptation structure. UniMem instead targets streaming settings by separating task identification from task execution through lightweight routing tokens and dedicated parametric memory blocks.

\subsection{Parameterized Memory for Streaming Task Internalization}
As LLM agents process streaming tasks, managing parametric memories without \emph{catastrophic forgetting} remains a critical challenge~\citep{bidaki2025onlinecontinuallearningsystematic,chen2026continuallearninglargelanguage}. While multitask tuning inherently causes knowledge entanglement, standard modular techniques like LoRA suffer from representation interference when new task patterns are sequentially absorbed. To mitigate this parameter collision, various architectures explore different dimensions of parameter isolation to partition task-specific knowledge. One prominent direction focuses on macro-level module orchestration, which includes imposing orthogonal subspace constraints on independent adapters in O-LoRA~\citep{wang2023orthogonalsubspacelearninglanguage} or employing dynamic routers for runtime adapter composition in L2R~\citep{araujo2024learningroutedynamicadapter}. Concurrently, another parallel line of research targets micro-level parameter manipulation, either by deploying LoRA-driven Mixture-of-Experts (MoE) networks like MixLoRA~\citep{li2024mixloraenhancinglargelanguage} to partition knowledge at the sub-task token level, or by encapsulating procedural memories into dedicated token slots as exemplified by Progressive Prompts~\citep{razdaibiedina2023progressivepromptscontinuallearning} and TOKMEM~\citep{wu2026tokmemonetokenproceduralmemory}.

\begin{figure*}[t]
\centering
\includegraphics[width=\textwidth]{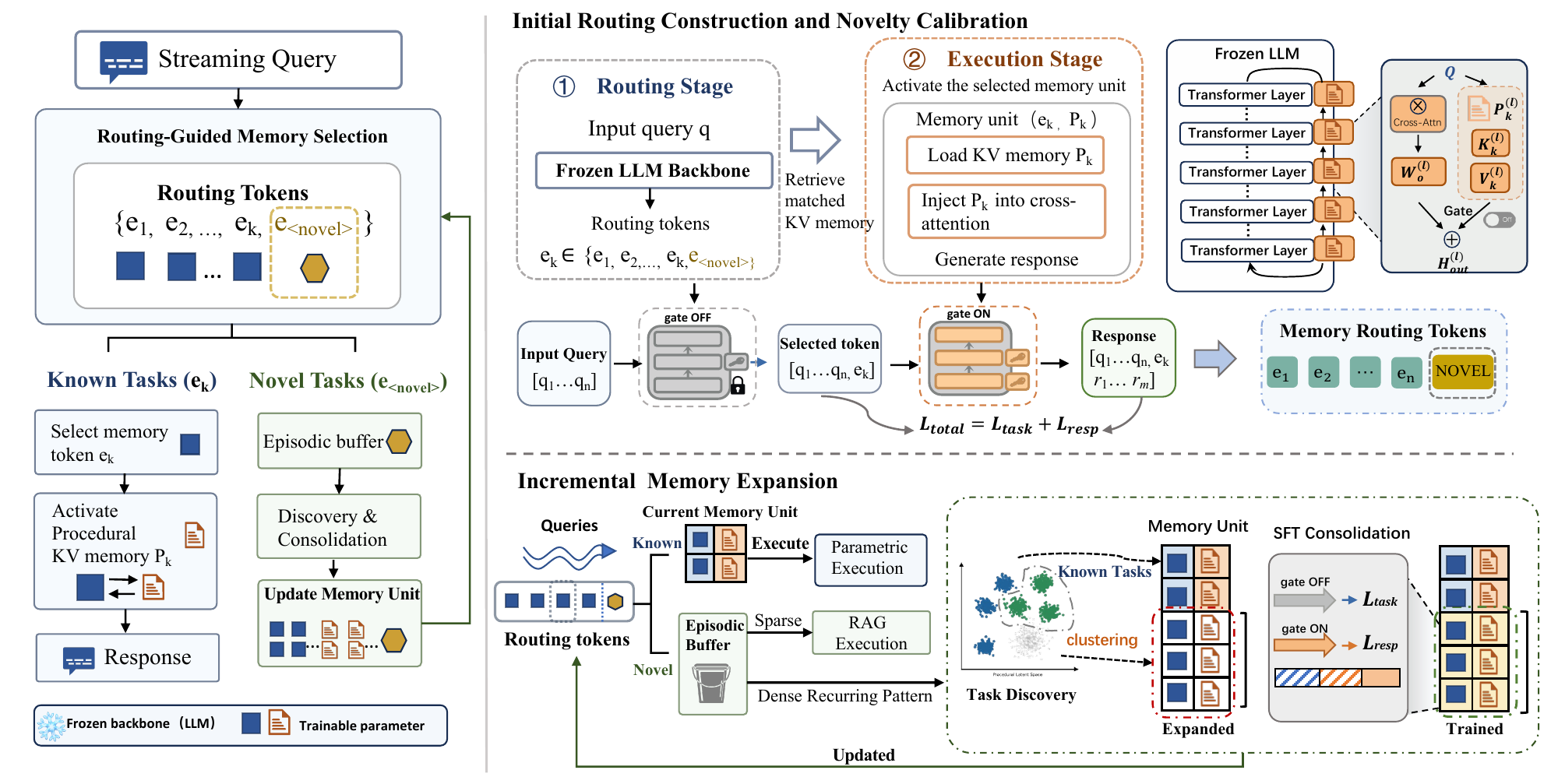}
\caption{Architecture of UniMem. The left panel shows the overall self-routing architecture for assigning incoming queries to known-task memory units or the novelty pathway. On the right, the upper part illustrates routing-guided memory selection and execution, while the lower part shows episodic buffering and parametric consolidation during memory expansion.}
\label{fig:overall_method}
\end{figure*}

However, most existing methods assume a predefined task set and rely on manually allocated or fixed-capacity parameter structures, making them difficult to adapt to boundary-agnostic streaming scenarios. UniMem addresses it through a self-routing episodic-to-parametric lifecycle, where sparse queries are handled by retrieval and recurring tasks are consolidated into parametric memory.

\section{Methodology}
\label{sec:method}

In this section, we formalize the proposed UniMem framework. We first establish the problem setting and the foundational token memory paradigm (Section~\ref{sec:preliminaries}). 
Then we introduce our self-routing memory architecture, which decouples lightweight routing tokens from high-capacity task-specific parametric memory 
blocks (Section~\ref{sec:architecture}). Finally, we detail the system's operational dynamics, including the two-stage episodic-to-parametric lifecycle from foundational memory initialization to autonomous task discovery and consolidation (Section~\ref{sec:incremental_expansion}).

\subsection{Problem Setting and Preliminaries}
\label{sec:preliminaries}
Our framework builds upon the procedural token memory paradigm \citep{wu2026tokmemonetokenproceduralmemory}, which compresses task-specific execution procedures into single token embeddings alongside a frozen pretrained LLM, denoted as $f_\theta$. For simplicity, we denote each reserved token by its trainable embedding $\mathbf{e}_k \in \mathbb{R}^d$, which represents the corresponding task procedure within the model. 
During training, the procedural token is inserted between the input query $q$ and the target response $r$, producing a serialized procedure-response sequence:
$x = (q_1, \dots, q_{|q|}, \; \mathbf{e}_k, \; r_1, \dots, r_{|r|}).
\label{eq:layout}$
% x = (q_1, \dots, q_{|q|}, \; \mathbf{e}_k, \; r_1, \dots, r_{|r|}).
% \label{eq:layout}
% \end{equation}

At inference, a query $q$ is routed by computing a distribution over the memory embeddings based on the final hidden state $\mathbf{h}_q$:
\begin{equation}
P(\mathbf{e}_k \mid q) =
\frac{\exp(\mathbf{h}_q^\top \mathbf{e}_k)}
{\sum_{j=1}^{K} \exp(\mathbf{h}_q^\top \mathbf{e}_j)} .
\label{eq:routing_prob}
\end{equation}

While effective as a compact task identifier, a single token embedding is insufficient to encode complex procedural logic.  We therefore retain the token embedding for routing and associate it with a separate high-capacity parametric memory block, as introduced in the next subsection.

\subsection{Self-Routing Memory Architecture}
\label{sec:architecture}

UniMem uses a self-routing memory selection mechanism that separates task identification from task execution. Given an incoming query, this mechanism decides whether to activate an existing parametric memory unit or route the query into the novelty-driven expansion pathway.

\paragraph{Routing Tokens and Memory Selection.}
Extending the single-token memory formulation, UniMem retains routing tokens as lightweight selectors rather than treating them as complete memories. Specifically, UniMem maintains a routing token matrix
$\mathbf{E} = [\mathbf{e}_1, \dots, \mathbf{e}_K, \mathbf{e}_{\texttt{NOVEL}}]$,
whose tokens fall into two functional classes:
\begin{itemize}
    \item \textbf{Known Task Tokens} ($\mathbf{e}_k$): Each known task token corresponds to a consolidated task and is associated with a parametric memory unit $\mathbf{u}_k = (\mathbf{e}_k, \mathbf{P}_k)$, where $\mathbf{P}_k$ stores task-specific execution parameters.
    \item \textbf{Novelty Sentinel Token} ($\mathbf{e}_{\texttt{NOVEL}}$): The novelty sentinel token is calibrated to capture queries that do not match existing task representations and route them to the episodic buffer.
\end{itemize}
Token selection follows a confidence-based rule: UniMem activates the memory unit of the top known token only when
its routing probability exceeds both the novelty-sentinel probability
and a threshold $\tau_{\mathrm{route}}$; otherwise, the query enters the novelty pathway (Appendix~\ref{app:routing_rule}). 
% Importantly, the memory block $\mathbf{P}_k$ is not activated during routing.

\paragraph{Known-Task Parametric Inference.}

Once a query is assigned to a known routing token $\mathbf{e}_k$, UniMem activates the associated memory unit $\mathbf{u}_k=(\mathbf{e}_k,\mathbf{P}_k)$ and enters the parametric execution phase. The memory block $\mathbf{P}_k$ is designed as a pluggable parameterization and can be instantiated with different PEFT-style modules. In this work, we instantiate $\mathbf{P}_k$ as a Procedural KV Memory composed of learnable, layer-wise key-value pairs $(\mathbf{K}_k^{(l)}, \mathbf{V}_k^{(l)})$.

During execution, the selected Procedural KV Memory contributes through the learned injection weights $\{g^{(l)}\}$. 
The resulting memory output is blended into the frozen self-attention output $\mathbf{H}_{\text{orig}}^{(l)}$:
\begin{equation}
\mathbf{H}_{\text{out}}^{(l)} = \mathbf{H}_{\text{orig}}^{(l)} + g^{(l)} \cdot \text{CrossAttn}\Big(\mathbf{Q}^{(l)}, \mathbf{K}_k^{(l)}, \mathbf{V}_k^{(l)}\Big),
\label{eq:cross_attn}
\end{equation}
where $\mathbf{Q}^{(l)}$ denotes the query states at layer $l$.

% If the query is routed to $\mathbf{e}_{\texttt{NOVEL}}$, no parametric memory block is invoked; the query is instead sent to the Episodic Buffer for retrieval-augmented execution and potential future consolidation, as described in Section~\ref{sec:incremental_expansion}.

\paragraph{Decoupled Optimization Objectives.}
To align training with the retrieve-then-execute workflow, we optimize routing and execution with two objectives.  The routing objective learns to select the appropriate routing token from the query alone, while the execution objective learns to generate the response using the selected parametric memory. Importantly, the execution loss is applied only to known-task samples, since samples routed to the novelty sentinel do not have an associated parametric memory block during calibration.

Let $z \in \{1,\dots,K,\texttt{NOVEL}\}$ denote the discrete routing label of an input query, and let $\Theta_k = \{\mathbf{P}_k, \{g^{(l)}\}\}$ denote the task-specific parameters activated when a known task $k$ is selected. The decomposed objective is:
\begin{equation}
\label{eq:decomposed_loss}
\begin{aligned}
\mathcal{L}_{\text{route}} 
&= -\log p(z \mid q_{1:n};\, \mathbf{E}), \\
\mathcal{L}_{\text{exec}}^{(k)}  
&= -\frac{1}{m}\sum_{t=1}^{m}
\log p(y_t \mid q_{1:n}, k, y_{<t};\, \Theta_k).
\end{aligned}
\end{equation}

During routing, all parametric memory blocks are inactive, so the routing objective updates only the routing tokens. The execution loss $\mathcal{L}_{\text{exec}}$ is applied only for known tasks, updating the selected parametric memory block $\mathbf{P}_k$ and its gates.

\subsection{Incremental Memory Expansion}
\label{sec:incremental_expansion}
UniMem starts with an initial set of known tasks and then expands memory from boundary-agnostic streaming samples through an episodic-to-parametric lifecycle. In Phase I, UniMem constructs the initial routing space and calibrates the novelty sentinel; in Phase II, it discovers recurring buffered patterns and consolidates qualified clusters into new parametric memory units.

\paragraph{Phase I: Routing Space Initialization.}
Phase I establishes a stable routing space for known tasks and calibrates a novelty sentinel for unseen queries. 
Given the initial known task set $\mathcal{T}^{(0)}$, UniMem first learns the routing tokens 
$\{\mathbf{e}_k\}_{k=1}^{K}$. After that, we compute the average routing norm
$
\bar{r} = \frac{1}{K}\sum_{k=1}^{K}\|\mathbf{e}_k\|,
$
which provides a fixed geometric scale for later memory expansion.

We then initialize the novelty sentinel $\mathbf{e}_{\texttt{NOVEL}}$ near the centroid of the known routing tokens. 
Specifically, a small Gaussian perturbation is added to the centroid direction, and the resulting vector is normalized to the same average norm:
\begin{equation}
\mathbf{e}_{\texttt{NOVEL}} =
\bar{r} \cdot
\frac{\bar{\mathbf{e}} + \boldsymbol{\epsilon}}
{\|\bar{\mathbf{e}} + \boldsymbol{\epsilon}\|},
\quad
\boldsymbol{\epsilon} \sim
\mathcal{N}\big(\mathbf{0}, (0.05\bar{r})^2\mathbf{I}\big),
\label{eq:novel_init}
\end{equation}
where $\bar{\mathbf{e}}$ denotes the mean direction of the known routing tokens. 
This initialization places the novelty sentinel within the same routing space as known tasks, allowing it to compete with existing routing tokens rather than behave as an outlier.

To calibrate the novelty sentinel, we reserve a subset of tasks from the initial training pool and treat them as pseudo-unseen tasks. These held-out tasks are used only for novelty calibration and are not allocated
known-task routing tokens.
The routing layer is then trained with a $(K+1)$-way cross-entropy objective over the known routing labels and the novelty label $\texttt{NOVEL}$. 
Samples from observed known tasks are assigned to their corresponding routing labels, whereas samples from the reserved pseudo-unseen tasks are assigned to $\texttt{NOVEL}$. 
This calibration teaches the routing layer to separate known task queries from queries that should enter the episodic buffer.

\paragraph{Phase II: Autonomous Task Discovery and Parametric Consolidation.}
During streaming deployment, queries routed to the novelty sentinel are stored in the Episodic Buffer and temporarily handled by retrieval-augmented execution. When the buffer reaches capacity $C$, UniMem triggers unsupervised task discovery. To avoid false consolidation, it uses a conservative clustering pipeline: NCD~\citep{cilibrasi2004clusteringcompression} measures task-pattern similarity, HDBSCAN~\citep{campello2013density} forms boundary-agnostic candidate clusters, and only clusters passing embedding-centroid outlier filtering and NCD-based cohesion checks are promoted. Details are provided in Appendix~\ref{app:ncd_streaming_clustering}.

\begin{table*}[t]
\centering
\small
\caption{Generative execution performance on the SNI streaming task sequence scaling from 10 to 100 tasks. Results are averaged over three runs using the same seed-42 data split. Results are presented as \textbf{Exact Match (\%) / ROUGE-L (\%)}. The best results in each column are highlighted in \textbf{bold}, and the second-best results are \underline{underlined}.}
\label{tab:sni_main_results}
\begin{tabular*}{\textwidth}{@{\extracolsep{\fill}} l cccc}
\toprule
\multirow{2}{*}{\textbf{Method}} & \multicolumn{3}{c}{\textbf{Number of Tasks ($T$)}} & \multirow{2}{*}{\textbf{Avg.}} \\
\cmidrule{2-4}
& \textbf{10} & \textbf{50} & \textbf{100} & \\
\midrule
\multicolumn{5}{c}{\textit{Backbone: LLaMA-3.2-3B}} \\
\midrule
Base          & 29.60 / 47.71 & 32.64 / 45.97 & 31.80 / 45.72 & 31.35 / 46.47 \\
RAG           & 37.00 / 58.43 & 37.22 / 51.79 & 35.21 / 49.45 & 36.48 / 53.22 \\
LoRA & \underline{38.40} / 60.36 & \underline{45.76} / \underline{57.91} & \underline{43.62} / \underline{56.67} & \underline{42.59} / \underline{58.31} \\
TokMem        & \underline{38.40} / \underline{61.80} & 41.06 / 54.67 & 40.54 / 55.07 & 40.00 / 57.18 \\
\textbf{UniMem (Ours)} & \textbf{51.60} / \textbf{67.01} & \textbf{47.22} / \textbf{59.53} & \textbf{47.56} / \textbf{60.58} & \textbf{48.79} / \textbf{62.37} \\
\midrule
\multicolumn{5}{c}{\textit{Backbone: LLaMA-3.1-8B}} \\
\midrule
Base          & 39.80 / 58.09 & 39.40 / 53.33 & 39.28 / 53.20 & 39.49 / 54.87 \\
RAG           & 45.20 / 64.83 & 42.61 / 59.12 & 41.56 / 56.65 & 43.12 / 60.20 \\
LoRA & 55.20 / \underline{73.80} & \underline{49.84} / 62.69 & \underline{49.39} / \underline{62.58} & \underline{51.48} / \underline{66.36} \\
TokMem        & \underline{56.20} / \textbf{74.08} & 49.27 / \underline{62.87} & 44.74 / 59.74 & 50.07 / 65.56 \\
\textbf{UniMem (Ours)} & \textbf{58.00} / 73.75 & \textbf{51.84} / \textbf{64.49} & \textbf{51.93} / \textbf{64.46} & \textbf{53.92} / \textbf{67.57} \\
\midrule
\multicolumn{5}{c}{\textit{Backbone: Qwen3-8B}} \\
\midrule
Base          & 40.80 / 53.10 & 45.84 / 57.28 & 44.36 / 57.72 & 43.67 / 56.03 \\
RAG           & \underline{41.80} / 60.35 & 42.86 / \underline{58.53} & 42.26 / 58.68 & 42.31 / 59.19 \\
LoRA & 44.80 / \underline{63.38} &\underline{47.51} / \underline{60.20} & \textbf{50.44} / \textbf{63.48} & \underline{47.58} / \underline{62.35} \\
TokMem        & 39.80 / 59.67 & 44.45 / 57.46 & 44.35 / 57.92 & 42.87 / 58.35 \\
\textbf{UniMem (Ours)} & \textbf{53.20} / \textbf{67.83} & \textbf{49.71} / \textbf{61.38} & \underline{49.94} / \underline{62.22} & \textbf{50.95} / \textbf{63.81} \\
\bottomrule
\end{tabular*}
\end{table*}

For each qualifying cluster, UniMem dynamically allocates a fresh routing token $\mathbf{e}_{\text{new}}$ paired with a new parametric memory block $\mathbf{P}_{\text{new}}$, instantiated in this work as Procedural KV Memory. The new memory units are then jointly optimized
on the clustered samples via supervised fine-tuning (SFT). Clusters that fail the quality gate remain in the Episodic Buffer and continue
to be served by RAG-based inference. This provides non-parametric coverage for
sparse or long-tail tasks until sufficient evidence accumulates for future
consolidation.

\section{Experiments}
\label{sec:experiments}
In this section, we evaluate the effectiveness of UniMem on streaming task sequences and focus on whether UniMem can maintain execution performance under boundary-agnostic streams while supporting stable memory expansion.

\subsection{Experimental Setup}
\label{subsec:setup}

\paragraph{Benchmarks and Streaming Protocols.}
We conduct evaluations on two representative benchmarks to assess UniMem across varying task complexities and scales. 
\textbf{(1) Super-Natural Instructions (SNI)}~\citep{wang2022supernaturalinstructionsgeneralizationdeclarativeinstructions}: To evaluate scalability across diverse tasks, we construct SNI streaming sequences with up to 100 tasks. We further simulate long-tail distributions by limiting a subset of tasks to only 10 training instances, causing them to remain episodic unless enough evidence accumulates for consolidation. Detailed construction procedures are provided in Appendix~\ref{app:streaming_protocol}. \textbf{(2) SuperGLUE Mixed Stream}~\citep{wang2020supergluestickierbenchmarkgeneralpurpose}: We select six complex reasoning tasks (BoolQ, CB, RTE, COPA, MultiRC, WiC) and blend them into a single, boundary-agnostic streaming task sequence.

%  This setup constructs a strictly \textbf{boundary-agnostic} environment, compelling the model to autonomously detect latent task patterns and route novel or uncertain queries without explicit task identifiers.
\paragraph{Implementation Details and Metrics.}
For SuperGLUE, we use LLaMA-1B and LLaMA-3B to evaluate reasoning streams under
smaller backbone scales. For the larger and more diverse SNI generative stream,
we further evaluate LLaMA-3.2-3B, LLaMA-3.1-8B, and Qwen3-8B to test scalability
across stronger backbones. Performance on SuperGLUE is quantified via Average Accuracy ($\mathcal{A}$), while SNI performance is measured using both Exact Match (EM) and ROUGE-L.
% as proxies for task-specific generation fidelity.

% The main results of UniMem on the SNI benchmark and the SuperGLUE mixed stream are summarized in Table~\ref{tab:sni_main_results} and Table~\ref{tab:main_results}.

\begin{table*}[t]
\centering
\small
\caption{Main results on the mixed SuperGLUE streaming task sequence. Performance is measured by final accuracy (\%) after processing the full boundary-agnostic stream.}
\label{tab:main_results}
% 使用 tabular* 和 \extracolsep{\fill} 完美自动撑满 \textwidth
\begin{tabular*}{\textwidth}{@{\extracolsep{\fill}} l ccccccc}
\toprule
\textbf{Method} & \textbf{BoolQ} & \textbf{CB} & \textbf{RTE} & \textbf{COPA} & \textbf{MultiRC} & \textbf{WiC} & \textbf{Avg. ($\mathcal{A}$)} \\
\midrule
\multicolumn{8}{c}{\textit{Backbone: LLaMA-1B-Instruct}} \\
\midrule
Base & 60.06 & 8.93 & 53.07 & 35.00 & 56.15 & 50.31 & 43.92 \\
RAG         & 47.95 & 1.79 & 46.93 & 24.00 & 52.37 & 27.90 & 33.49 \\
LoRA & 76.97 & 51.79 & 68.23 & 68.00 & 72.73 & 51.72 & 64.91 \\
TOKMEM & 72.80 & 1.80 & 68.60 & 3.00 & 69.90 & 49.50 & 44.27 \\
\textbf{UniMem (Ours)} & \textbf{81.96} & \textbf{69.64} & \textbf{80.14} & \textbf{78.00} & \textbf{75.91} & \textbf{57.68} & \textbf{73.89} \\
\midrule
\multicolumn{8}{c}{\textit{Backbone: LLaMA-3B-Instruct}} \\
\midrule
Base     & 81.19 & 60.71 & 72.92 & 39.00 & 70.77 & 50.31 & 62.48 \\
RAG         & 75.87 & 78.57 & 74.37 & 81.00 & 68.03 & 52.66 & 71.75 \\
LoRA & 86.54 & 75.00 & 82.67 & 83.00 & 85.21 & 52.51 & 77.49 \\
TOKMEM & 84.40 & 64.30 & 80.90 & \textbf{94.00} & 84.00 & 66.30 & 78.98 \\
\textbf{UniMem (Ours)} & \textbf{88.07} & \textbf{87.50} & \textbf{88.09} & 92.00 & \textbf{86.55} & \textbf{73.51} & \textbf{85.95} \\
\bottomrule
\end{tabular*}
\end{table*}

\begin{figure}[t]
\centering
\includegraphics[width=\columnwidth]{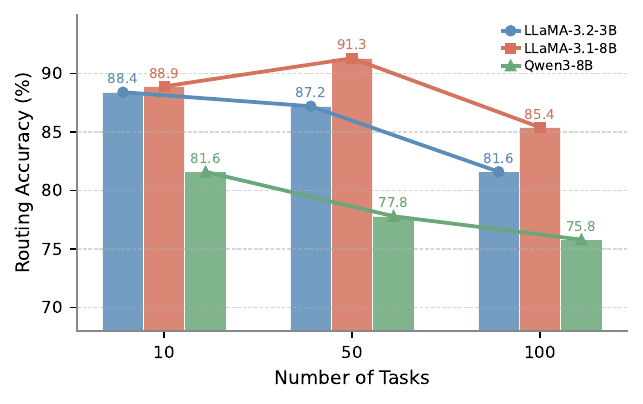}
\caption{
Routing token accuracy on SNI streaming task sequences under increasing task scales. 
}
\label{fig:routing_accuracy}
\end{figure}

\paragraph{Baselines.}
We compare UniMem with four representative baselines covering non-memory, external-memory, parametric-memory, and token-memory paradigms: \textbf{Base (Zero-Shot)}, the vanilla pre-trained model without task-specific adaptation; \textbf{Standard RAG}, which retrieves the $k$-nearest historical examples as in-context evidence; \textbf{Replay LoRA}, a rehearsal-based parametric baseline that interleaves a fixed proportion of previous samples (10\%) into each training block; and \textbf{Adapted TOKMEM~\citep{wu2026tokmemonetokenproceduralmemory}}, a token-centric procedural memory baseline adapted to operate
without explicit task labels in our streaming protocol. Detailed baseline implementation settings are provided in Appendix~\ref{app:baseline_details}.

\subsection{Main Results and Analysis}\label{sec:results}

\paragraph{Performance on SNI Streaming Tasks.}
\label{subsec:sni_analysis}
Table~\ref{tab:sni_main_results} presents execution performance on the SNI streaming task sequence. UniMem achieves the best average scores across three backbone models, suggesting that UniMem can sustain strong task execution during long-term memory expansion.

The comparison with TOKMEM and Replay LoRA highlights the benefit of separating routing from execution. TOKMEM compresses task-specific behavior into a single token, whereas UniMem uses routing tokens only for task selection and stores execution behavior in separate parametric memory blocks. This design better supports diverse task procedures: on LLaMA-3.2-3B with 100 tasks, UniMem achieves 47.56\% EM, compared with 40.54\% for TOKMEM.

Replay LoRA is a strong parametric baseline, but its fixed parameter space can entangle heterogeneous task patterns. In contrast, UniMem consolidates recurring patterns into separate parametric memory units while keeping sparse tasks retrieval-based. As shown in Table~\ref{tab:sni_main_results}, UniMem improves average EM over Replay LoRA by +6.20 and +2.44 points on LLaMA-3.2-3B and LLaMA-3.1-8B, respectively, with higher ROUGE-L scores.

We further evaluate routing token accuracy on the SNI streaming tasks. As shown in Figure~\ref{fig:routing_accuracy}, UniMem preserves reliable routing performance across task scales, suggesting that the routing layer can consistently identify the appropriate parametric memory unit. The LLaMA backbones achieve particularly strong accuracy, with LLaMA-3.1-8B remaining above 85\% at Test-100. Although accuracy gradually decreases as semantically similar instructions and overlapping task patterns increase routing difficulty, the overall results indicate that UniMem 
maintains sufficiently 
discriminative routing tokens for 
long-horizon memory selection.

\begin{figure}[t]
\centering
\includegraphics[width=\columnwidth]{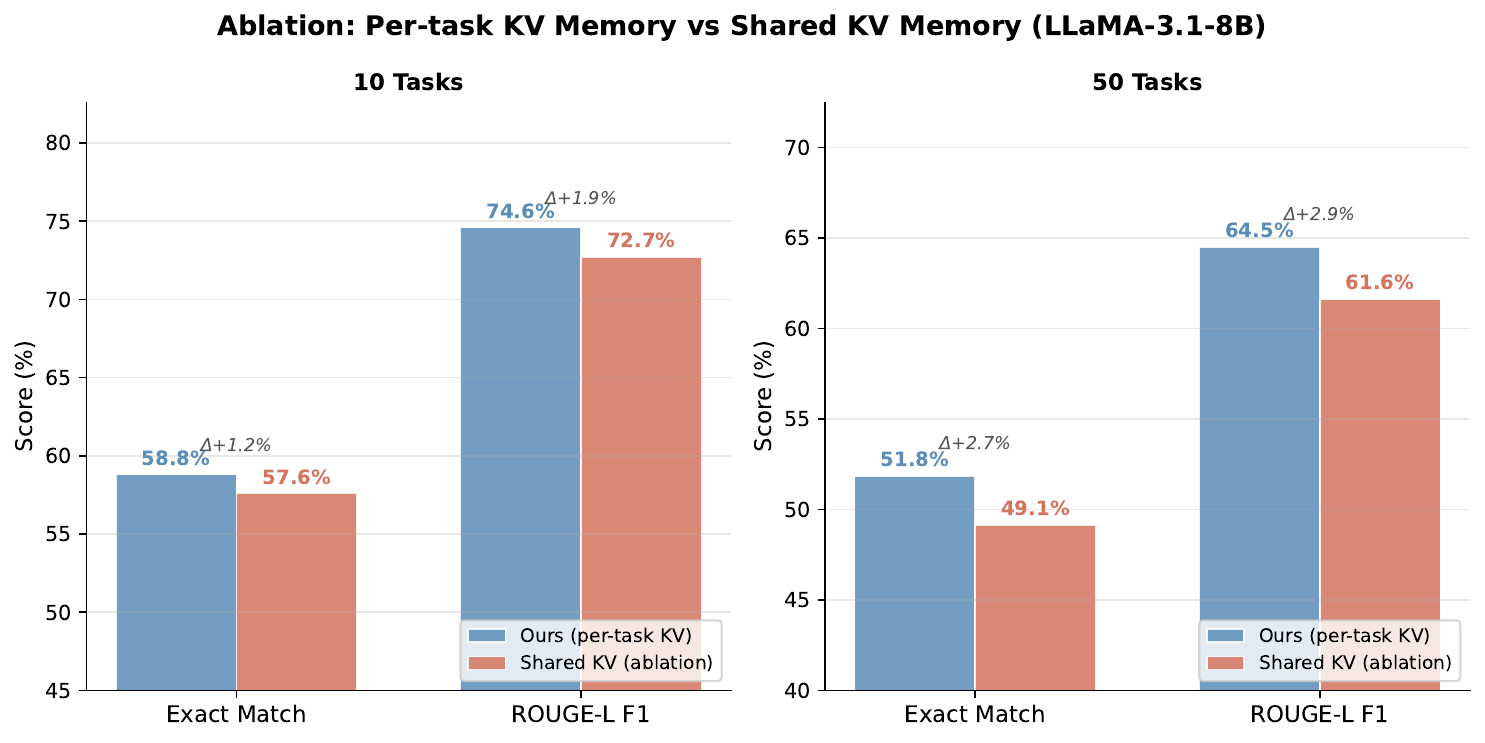}
\caption{Ablation study of per-task vs. shared KV prefixes on LLaMA-3.1-8B.}
\label{fig:ablation_kv}
\end{figure}

% \subsubsection{Performance on Mixed Task Streams}
% \label{sec:superglue_results}

% Table~\ref{tab:main_results} presents the evaluation results on the unstructured, unlabeled SuperGLUE Mixed Stream. Operating in a boundary-agnostic environment, \textbf{UniMem} consistently outperforms the compared baselines in overall average accuracy ($\mathcal{A}$) across both backbone scales.

\paragraph{Performance on SuperGLUE Streams.}
\label{sec:superglue_results}

Table~\ref{tab:main_results} reports results on the boundary-agnostic and boundary-agnostic SuperGLUE Mixed Stream. UniMem achieves the highest average accuracy ($\mathcal{A}$) across both backbone scales, suggesting that it can handle implicit task transitions more effectively than baselines relying on a single shared adaptation space or token-centric memory representations. In particular, UniMem performs strongly on challenging tasks such as COPA and achieves the best average performance on the LLaMA-3B backbone.

\begin{table}[ht]
\centering
\small
\caption{Ablation study on core components of \textbf{UniMem} across different streaming task scales. Results are reported as \textbf{EM (\%) / ROUGE-L (\%)}.}
\label{tab:ablation_study}
\begin{tabular}{l cc cc}
\toprule
\multirow{2}{*}{\textbf{Method}} & \multicolumn{2}{c}{\textbf{Test-10}} & \multicolumn{2}{c}{\textbf{Test-50}} \\
\cmidrule(lr){2-3} \cmidrule(lr){4-5}
& \textbf{EM} & \textbf{R-L} & \textbf{EM} & \textbf{R-L} \\
\midrule
\multicolumn{5}{c}{\textit{Backbone: LLaMA-3.2-3B}} \\
\midrule
UniMem (full)       & \textbf{51.60} & \textbf{67.01} & \textbf{47.22} & \textbf{59.53} \\
~~$w/o$ RAG cache & ---            & ---            & 46.45          & 58.97          \\
~~$w/o$ KV gate   & 35.80          & 56.43          & 31.22          & 45.33          \\
\midrule
\multicolumn{5}{c}{\textit{Backbone: LLaMA-3.1-8B}} \\
\midrule
UniMem (full)       & \textbf{58.00} & \textbf{73.75} & \textbf{51.84} & \textbf{64.49} \\
~~$w/o$ RAG cache & ---            & ---            & 50.90          & 63.10          \\
~~$w/o$ KV gate   & 37.40          & 57.10          & 30.16          & 42.05          \\
\bottomrule
\end{tabular}
\end{table}

\begin{figure}[t]
\centering
\includegraphics[width=\columnwidth]{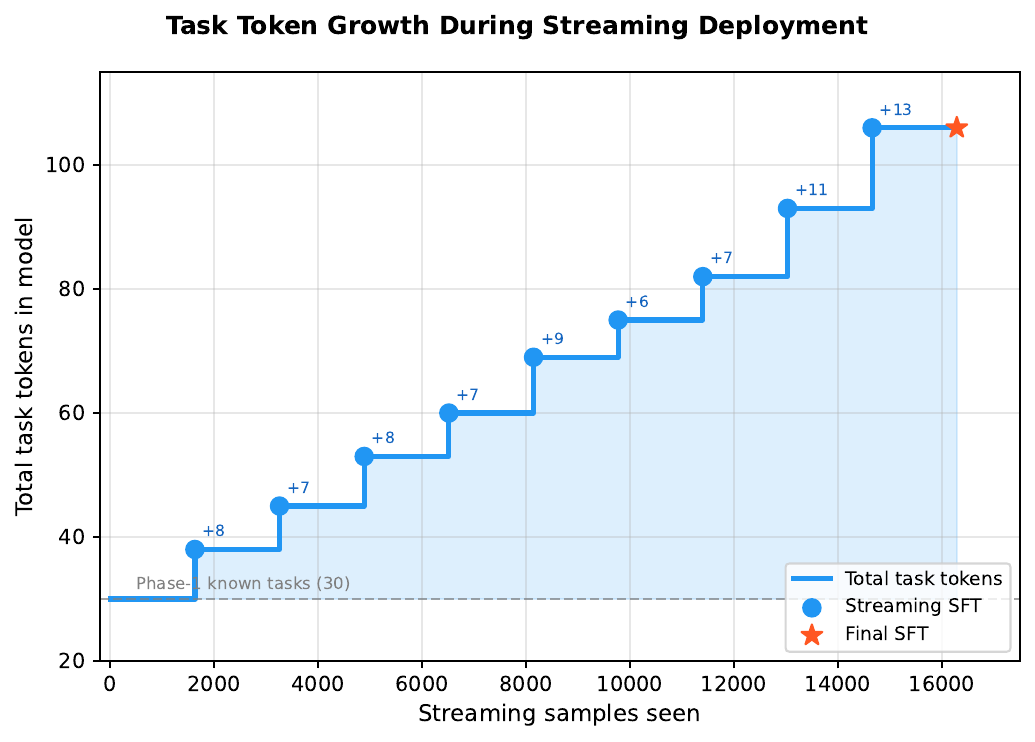}
\caption{Cumulative memory growth during streaming deployment.}
\label{fig:growth_curve}
\end{figure}

\subsection{Ablation and Qualitative Analysis}\label{subsec:ablation}

\textbf{Effectiveness of Isolated Procedural KV Memory.}
We compare UniMem with a Shared KV Memory baseline on LLaMA-3.1-8B to assess the value of task-specific memory blocks. As shown in Figure~\ref{fig:ablation_kv}, UniMem outperforms the shared variant in both 10-task and 50-task settings. The EM advantage increases from +1.2\% to +2.7\% as the task sequence grows, suggesting that shared parameters become more vulnerable to interference under greater task diversity. In contrast, task-specific KV memory helps preserve stable execution during streaming expansion.

\textbf{Impact of Episodic Buffer Retrieval.}
The ablation of Episodic Buffer retrieval ($w/o$ Episodic Buffer RAG) is omitted in the 10-task setting, where only one sparse query is observed and retrieval-augmented execution becomes highly volatile. In the 50-task setting, sparse and long-tail queries appear more frequently, and disabling Episodic Buffer retrieval leads to performance drop. This confirms that the Episodic Buffer provides useful non-parametric coverage for queries that do not yet justify parametric consolidation.

\textbf{Criticality of the Gating Mechanism.}
Disabling the gating mechanism ($w/o$ KV gate) causes substantial performance degradation across all settings. On the LLaMA-3.2-3B backbone, the EM score drops sharply in both the 10-task and 50-task settings, with a similar decline observed on LLaMA-3.1-8B. This confirms that learned gates are crucial for regulating the injection of task-specific Procedural KV Memory into the frozen backbone; without them, heterogeneous procedural signals interfere more strongly during execution.

\textbf{Dynamics of Autonomous Task Discovery.}
Figure~\ref{fig:growth_curve} tracks memory growth during streaming. Starting from 30 foundational tasks, UniMem allocates new parametric units only when buffered queries form recurring clusters. Across 16,000 samples covering 100 novel tasks, 76 new parametric units are created, while 10 sparse tasks remain in the episodic buffer due to insufficient evidence, and several semantically similar tasks merge into known units. This demonstrates that UniMem expands memory according to recurring task structure rather than allocating one unit per novel task.

\section{Conclusion}

In this paper, we introduced \textbf{UniMem}, a unified framework that bridges episodic retrieval and parametric consolidation for open-world memory expansion under streaming tasks. UniMem separates lightweight routing tokens from task-specific procedural memory, enabling LLM agents to handle heterogeneous task streams while keeping the backbone frozen. Through a novelty-guided lifecycle, sparse or uncertain queries are retained in an Episodic Buffer, whereas recurring workflows are selectively consolidated into persistent parametric memory. Experiments on long-horizon streaming task sequences show that UniMem maintains robust execution fidelity and outperforms token-centric memory methods and replay-based baselines. These results suggest a scalable direction for building self-evolving agents with expandable parametric memory.


\section*{Limitations}

Although UniMem shows promising results for self-routing memory expansion under streaming tasks, two directions remain open. First, UniMem adopts a conservative consolidation strategy, keeping sparse or uncertain patterns in the Episodic Buffer until sufficient evidence is observed. While this avoids premature parameter allocation, more adaptive consolidation criteria may further improve memory utilization in dynamic streams.

Second, this work instantiates parametric memory with Procedural KV Memory and keeps the backbone frozen. Other parameter-efficient memory modules may offer different trade-offs in capacity, efficiency, and transferability, which we leave for future exploration.

\bibliography{custom}

\appendix
% ============ 附录部分 ============

\section{Routing Selection Mechanism}
\label{app:routing_rule}

This appendix describes the inference-time routing rule used by
UniMem. The rule is designed to support two complementary novelty
signals: explicit novelty recognition by the learned novelty
token, and low-confidence rejection when no known task can
confidently explain the input.

Given an input query $\mathbf{x}$, the frozen backbone produces the
last-token hidden state $\mathbf{h} \in \mathbb{R}^{d}$. UniMem then
computes a routing distribution over $T+1$ candidates, consisting
of $T$ known-task routing tokens $\{e_1,\ldots,e_T\}$ and one novelty
sentinel $e_{\mathrm{NOV}}$. We index the novelty sentinel by
$i=0$:
\begin{equation}
p_i =
\frac{
\exp(\mathbf{h}^{\top} e_i / \sqrt{d})
}{
\sum_{j=0}^{T}
\exp(\mathbf{h}^{\top} e_j / \sqrt{d})
},
\quad i \in \{0,1,\ldots,T\}.
\end{equation}

Let $p_0$ denote the probability assigned to the novelty sentinel.
We define the strongest known-task candidate as
\begin{equation}
i^\star = \arg\max_{i \in \{1,\ldots,T\}} p_i,
\quad
p^\star = p_{i^\star}.
\end{equation}
Here, $i^\star$ is the best known-task routing token and $p^\star$
is its confidence.

UniMem activates a known parametric memory unit only when two
conditions are simultaneously satisfied:
\begin{equation}
p^\star > p_0
\quad \text{and} \quad
(\ p^\star \geq \tau_{\mathrm{route}}).
\end{equation}
Otherwise, the query is routed to the novelty pathway. Equivalently,
the final routing decision is
\begin{equation}
\hat{y} =
\begin{cases}
i^\star,
& \text{if } p^\star > p_0
\text{ and } ( p^\star \geq \tau_{\mathrm{route}}), \\[4pt]
e_{\mathrm{NOV}},
& \text{otherwise}.
\end{cases}
\end{equation}

This rule captures two ways in which a query can be treated as
novel. First, if the novelty sentinel itself receives higher
probability than every known-task token, the query is explicitly
recognized as novel. Second, even if a known-task token has the
largest probability among known tasks, UniMem rejects the known
memory activation when the confidence $p^\star$ is below the
threshold $\tau_{\mathrm{route}}$. This prevents low-confidence queries from
activating an incorrect Procedural KV Memory block.

When $\tau_{\mathrm{route}}=0$, the rule reduces to sentinel-based routing between
the strongest known-task token and the novelty sentinel. When
$\tau_{\mathrm{route}}>0$, the threshold introduces an additional conservative
rejection mechanism. In our experiments, $\tau_{\mathrm{route}}$ is set as $0.7$.

We use this max-known-probability gate instead of entropy or margin.
Entropy can be sensitive to the number of weak competing routing
tokens, while margin-based rejection introduces an additional
criterion between the top two candidates. The scalar $p^\star$
directly measures whether any known memory unit claims the query
with sufficient confidence, making the novelty decision simple and
interpretable.

\section{Procedural KV Memory: Implementation Details}
\label{app:kv_memory_details}

\paragraph{Cross-Attention Mechanics.}
The Procedural KV Memory in Eq.~\ref{eq:cross_attn} is implemented as a set of task-specific key-value slots at each Transformer layer. For a selected known task $k$, the current hidden states at layer $l$ are projected into query states $\mathbf{Q}^{(l)}$ using the frozen query projection $\mathbf{W}_Q^{(l)}$. We do not apply rotary position encoding (RoPE) to the Procedural KV Memory, since these key-value slots encode task-level procedural semantics rather than position-dependent token interactions. No causal mask is applied to this cross-attention, allowing the selected procedural memory to be globally accessible to every sequence position.

When the number of memory slots is $p{=}1$, the cross-attention reduces to a simple and interpretable form. Since there is only one key, the attention weight over the memory slot is always 1, and every position receives the same task-specific value vector after output projection. In this case, the Procedural KV Memory behaves as a layer-wise task-specific residual signal added to the frozen backbone representation.

\paragraph{Grouped-Query Attention (GQA) Handling.}
For models using grouped-query attention (GQA), such as Llama-3.2-3B with $H_q{=}32$ query heads and $H_{\text{kv}}{=}8$ key-value heads, the Procedural KV Memory is parameterized over $H_{\text{kv}}$ heads to reduce parameter cost. Before computing cross-attention, the stored key-value pairs are repeated $H_q / H_{\text{kv}}$ times to match the number of query heads. For Llama-3.2-3B, this corresponds to a repetition factor of $4$.

\paragraph{Gate Parameterization.}
Each layer is equipped with a learnable scalar gate that controls the contribution of the selected Procedural KV Memory. We use a clamped parameterization
\begin{equation}
g^{(l)} = \mathrm{clamp}(\tilde{g}^{(l)}, 0, g_{\max}),
\end{equation}
rather than a sigmoid gate. This choice provides a directly interpretable injection strength and avoids vanishing gradients near sigmoid saturation. We initialize $\tilde{g}^{(l)}{=}0.01$, so that the model starts close to the frozen backbone behavior and the Procedural KV Memory contributes only weakly at the beginning of training.

\paragraph{Parameter Budget.}
For a model with $L$ layers, $H_{\text{kv}}$ key-value heads, head dimension $d_h$, hidden dimension $d$, and $p$ memory slots per layer, each task adds trainable parameters.
\begin{equation}
    \underbrace{d}_{\text{routing token}}
    +
    \underbrace{2 L H_{\text{kv}} d_h p}_{\text{Procedural KV Memory}}
    +
    \underbrace{L}_{\text{gates}}
\end{equation}
For Llama-3.2-3B ($L{=}28$, $H_{\text{kv}}{=}8$, $d_h{=}128$, $d{=}3072$, $p{=}1$), this yields
\[
3{,}072 + 57{,}344 + 28 = 60{,}444
\]
parameters per task, which is several orders of magnitude smaller than the frozen 3B backbone.

\paragraph{Training Hyperparameters}
We summarize the main training and system hyperparameters used in the SNI streaming experiments. 

\begin{table}[h]
\centering
\small
\begin{tabular}{ll}
\toprule
\textbf{Hyperparameter} & \textbf{Value} \\
\midrule
Batch size & 2 \\
Gradient accumulation & 2 \\
Effective batch size & 4 \\
Maximum sequence length & 2560 \\
Precision & bfloat16 \\
Epochs per training round & 1 \\
Routing-token learning rate & $1 \times 10^{-3}$ \\
Procedural KV Memory learning rate & $5 \times 10^{-3}$ \\
\bottomrule
\end{tabular}
\caption{Training hyperparameters shared across the SNI streaming experiments.}
\label{tab:training_hyperparameters}
\end{table}
The Procedural KV Memory uses a larger learning rate of $5 \times 10^{-3}$, while the routing tokens are trained with a learning rate of $1 \times 10^{-3}$.

\section{Streaming Protocol Details}
\label{app:streaming_protocol}

\subsection{Artifact use.}
We use Super-NaturalInstructions and SuperGLUE as public benchmark datasets for research evaluation. We follow the original licenses and terms of use of all datasets and pretrained models used in this work, including the LLaMA and Qwen model families. We do not redistribute the original datasets or pretrained model weights. The constructed streaming splits are used only for controlled research evaluation.

\subsection{Task Pool Construction}

We construct the streaming task pool from the full English task set of Natural Instructions 2.8.
To ensure reproducibility, tasks are first sorted lexicographically and then shuffled with a fixed random seed, which is set to 42 by default.
From the shuffled list, we select $N_{\text{total}}$ tasks and partition them into four disjoint subsets: $T_{\text{stream}}$, $T_{\text{known}}$, $T_{\text{calib}}$, and $T_{\text{held}}$, corresponding to the unknown task stream, initial known tasks, novelty-calibration tasks, and unused held-out tasks, respectively.

The sizes of the first three subsets are
$N_{\mathrm{stream}}$, $N_{\mathrm{known}}$, and $N_{\mathrm{calib}}$.
Here, $N_{\mathrm{known}}$, $N_{\mathrm{calib}}$, and $N_{\mathrm{stream}}$ denote the number of initial known tasks, novelty-calibration tasks, and unknown streaming tasks, respectively.
In our experiments, we set $N_{\mathrm{known}}=20$ and $N_{\mathrm{calib}}=20$ for routing-space initialization and novelty calibration.
The unknown streaming tasks are kept strictly disjoint from the novelty-calibration tasks and unused held-out tasks to avoid label conflicts.
In particular, a streaming task should not appear as a novelty-calibration task, otherwise it would be treated both as a novelty target during calibration and later as a specific discovered cluster during streaming expansion.

\subsection{Per-Task Sample Split}

For each unknown streaming task, samples are sorted by a deterministic hash
\[
h(x)=\mathrm{MD5}(\mathrm{id}\,||\,\mathrm{input}\,||\,\mathrm{output}),
\]
where $||$ denotes string concatenation.
This deterministic ordering ensures that the same examples are selected across different runs.
The sorted samples are then split into test, train, and validation subsets:
\[
\underbrace{x_{1:N_{\mathrm{test}}}}_{\text{test}}
\quad
\underbrace{x_{N_{\mathrm{test}}+1:N_{\mathrm{test}}+N_{\mathrm{train}}}}_{\text{train}}
\quad
\underbrace{x_{N_{\mathrm{test}}+N_{\mathrm{train}}+1:\cdots}}_{\text{validation}} .
\]
Test samples are always drawn from the hash-sorted head, so the evaluation set remains fixed regardless of how many training samples are used.
Unless otherwise specified, we use $N_{\mathrm{train}}=200$ and $N_{\mathrm{test}}=50$ per task.

\subsection{Sparse Task Simulation}

To simulate long-tail task distributions in open-world environments, a fraction $r_{\mathrm{sparse}}$ of unknown streaming tasks is randomly designated as sparse tasks using an independent random generator with seed $42+9999$.
By default, $r_{\mathrm{sparse}}=10\%$.
Each sparse task is allocated only $N_{\mathrm{sparse}}=10$ training samples, which is far below the minimum cluster size $S_{\min}=50$ used by HDBSCAN.
As a result, sparse tasks cannot form valid clusters for parametric consolidation and are instead served through the episodic RAG fallback during inference.
This design tests whether UniMem can avoid premature parameter allocation for insufficiently supported task patterns.

\subsection{Stream Ordering}

Training samples from unknown streaming tasks are presented to the model as an boundary-agnostic stream.
We support two stream ordering modes in the implementation.

\textbf{Global shuffle.}
When this option is set to 0, all training samples from unknown streaming tasks are globally shuffled.
This is the default setting used in our experiments.
It simulates a mixed deployment stream in which task identity is unknown and samples from different tasks arrive interleaved.

\textbf{Group-wise stream.}
When this option is set to $G$, tasks are divided into consecutive groups of size $G$.
Within each group, samples from all $G$ tasks are pooled and locally shuffled, and groups arrive sequentially.
This setting preserves temporal locality among related tasks while still hiding task labels within each group.

In the main SNI experiments, we use the group-wise stream setting to simulate a multi-task deployment scenario where task groups arrive sequentially over time. Within each group, samples from multiple tasks are mixed without task labels or explicit task boundaries, requiring the model to perform routing and memory expansion under local task heterogeneity. For SuperGLUE, since the task set is relatively small, we instead use the global-shuffle setting, where samples from all tasks are fully mixed into a single boundary-agnostic stream.

\section{Baseline Implementation Details}
\label{app:baseline_details}

This appendix provides the implementation details of the baselines used in our experiments. 
For a controlled comparison, all trainable baselines are constructed from the same streaming training data and evaluated on the same held-out test split. Unless otherwise specified, all methods use greedy decoding during inference.

\subsection{Zero-Shot Baseline}

The zero-shot baseline directly applies the frozen pretrained model without any task-specific adaptation. In our experiments, we use the same backbone models as in the main evaluation. 

At inference time, each test sample from the   task set is formatted as a standard chat prompt, where the task instruction is prepended to the input query. The model then generates the response with greedy decoding, i.e., \texttt{do\_sample=False}. No task-specific parameters are introduced or updated. This baseline measures the inherent instruction-following and task generalization ability of the frozen pretrained model.

\subsection{Retrieval-Augmented Generation Baseline}

The retrieval-augmented generation (RAG) baseline augments zero-shot inference with retrieved in-context demonstrations. Its retrieval corpus is constructed from the training set, which is the same data observed by the streaming memory baselines. This ensures that RAG and the trainable streaming methods have access to the same pool of task evidence.

For each test query, we form the retrieval query by concatenating the task instruction and the input query. We then encode the retrieval query and all candidate training samples with Qwen3-Embedding-0.6B. Similarity is computed by cosine similarity over $\ell_2$-normalized embeddings. The top-$k$ most similar training samples are retrieved, where we set $k=3$ in all experiments.

The retrieved samples are formatted as few-shot demonstrations and prepended to the input prompt before generation. To control the context length, we impose a character budget of 4,000 characters for the retrieved demonstrations. The final prompt is decoded greedily without any gradient update. Thus, this baseline evaluates whether non-parametric retrieval alone can provide sufficient task adaptation under the same streaming data source.

\subsection{LoRA Fine-Tuning Baseline}

The LoRA baseline applies parameter-efficient fine-tuning to the   training data. The pretrained backbone remains frozen, and only the LoRA adapter parameters are updated. The training input is formatted with the same chat template used by the memory-based methods, where the task instruction is concatenated into the user turn. The loss is computed only on the assistant response tokens, while the instruction and query tokens are masked with the ignore label.

The LoRA configuration is shown in Table~\ref{tab:lora_config}.

\begin{table}[h]
\centering
\begin{tabular}{ll}
\toprule
\textbf{Hyperparameter} & \textbf{Value} \\
\midrule
LoRA rank $r$ & 8 \\
LoRA $\alpha$ & 32 \\
Dropout & 0.1 \\
Target modules & \texttt{q\_proj}, \texttt{v\_proj} \\
Learning rate & $5 \times 10^{-5}$ \\
Batch size & 1 \\
Gradient accumulation & 4 \\
Epochs & 1 \\
Maximum sequence length & 2560 \\
Precision & bfloat16 \\
\bottomrule
\end{tabular}
\caption{Hyperparameter configuration of the LoRA fine-tuning baseline.}
\label{tab:lora_config}
\end{table}

To ensure data alignment, the LoRA baseline uses the same data preparation procedure as the streaming memory methods. Specifically, it calls the same \texttt{prepare\_incremental\_data} routine with identical arguments, including the random seed, the number of   tasks, and the per-task training size. Therefore, LoRA observes exactly the same   training samples and is evaluated on the same held-out test set.

\subsection{LoRA with Continual Replay}

We further evaluate a continual replay variant of LoRA to mitigate catastrophic forgetting under sequential training. This baseline follows the same LoRA configuration as described above, but introduces block-wise cumulative replay during   adaptation.

The   tasks are divided into consecutive blocks of $B=10$ tasks. When training on the first block, no replay is used because no previous samples are available. For each subsequent block $b>0$, a fraction $\rho=0.1$ of the current block's training samples is replaced by samples uniformly drawn without replacement from the cumulative pool of all previous blocks. The number of replay samples is computed as
\begin{equation}
n_{\mathrm{replay}} = \left\lfloor n_{\mathrm{current}} \times \rho \right\rfloor ,
\end{equation}
where $n_{\mathrm{current}}$ denotes the number of samples in the current training block.

Replay samples are interleaved into the current training sequence at approximately uniform intervals. Specifically, one replay sample is inserted every$
\left\lfloor \frac{n_{\mathrm{current}}}{n_{\mathrm{replay}}+1} \right\rfloor
$
training steps. After training on block $b$, the samples from this block are added to the cumulative replay pool, allowing later blocks to replay from all earlier tasks.

We preserve the block-wise training order and do not globally shuffle the   stream, so that temporal task locality is maintained. All other training hyperparameters are identical to the plain LoRA baseline. In the main experiments, this replay-enhanced variant is reported as the Replay LoRA baseline.

\subsection{Adapted TOKMEM Baseline}
\label{app:tokmem_baseline}

We include an adapted TOKMEM baseline to provide a direct architectural comparison with UniMem. 
TOKMEM represents each task procedure with a single trainable memory token. 
In our implementation, we adapt TOKMEM to the same streaming protocol as UniMem, so that the two methods differ only in whether a routed memory token is paired with an additional high-capacity parametric memory block.

Specifically, the adapted TOKMEM baseline uses the same routing-space initialization, novelty-sentinel calibration, episodic buffering, task discovery, and consolidation schedule as UniMem. 
For each consolidated task or discovered cluster, TOKMEM allocates a trainable procedural token $e_k$. 
During inference, the routing rule selects the most appropriate token in the same way as UniMem. 
If the query is assigned to a known task token, the selected token $e_k$ is inserted into the input sequence and used directly as the task-specific procedural memory. 
If the query is routed to the novelty sentinel, it is handled by the same episodic retrieval fallback as in UniMem.

The key difference is that TOKMEM does not introduce an additional Procedural KV Memory block. 
In other words, each memory unit in TOKMEM contains only
\begin{equation}
u_k^{\mathrm{TOKMEM}} = \{ e_k \},
\end{equation}
whereas each UniMem memory unit contains both a routing token and an execution memory block:
\begin{equation}
u_k^{\mathrm{UniMem}} = \{ e_k, P_k \}.
\end{equation}
Therefore, the selected token in TOKMEM must simultaneously serve as the task identifier and the execution memory. 
By contrast, UniMem uses $e_k$ only as a lightweight selector and stores task-specific execution behavior in the associated Procedural KV Memory $P_k$.

Training follows the same response-supervised objective and data split as UniMem. 
For known or consolidated tasks, the loss is computed only on the assistant response tokens, while the instruction, query, and memory-token positions are masked. 
Only the memory token parameters are updated for TOKMEM, and the frozen backbone remains unchanged. 
No layer-wise key-value slots, cross-attention injection, or memory gates are used in this baseline.

This adapted TOKMEM baseline isolates the effect of decoupling routing from execution. 
If a single procedural token is sufficient to encode the task-specific behavior, TOKMEM should perform similarly to UniMem. 
The observed performance gap therefore reflects the benefit of augmenting lightweight routing tokens with dedicated Procedural KV Memory for high-capacity execution.

\section{NCD-Based Streaming Clustering}
\label{app:ncd_streaming_clustering}

This appendix provides the implementation details of the
unsupervised task discovery module. The goal of
this module is to identify recurring latent task patterns from
boundary-agnostic samples routed to the novelty sentinel. Since no task
labels or explicit task boundaries are available during streaming
deployment, UniMem performs clustering directly over the episodic
cache and promotes only high-quality clusters into parametric
procedural memory.

\subsection{Overview}

Let $\mathcal{B}$ denote the episodic cache that stores samples routed
to the novelty sentinel $e_{\mathrm{NOVEL}}$. Incoming samples are
appended to $\mathcal{B}$ until the cache size reaches a predefined
capacity $C$, where we set $C=1600$ by default. Once
$|\mathcal{B}| \geq C$, UniMem triggers one round of structural
discovery.

Each clustering round consists of three stages:
\begin{enumerate}
    \item computing sample-level procedural distances with
    Normalized Compression Distance (NCD);
    \item applying HDBSCAN to discover density-based candidate
    clusters without task labels;
    \item filtering noisy members and low-quality clusters before
    using the remaining clusters for procedural consolidation.
\end{enumerate}

Only clusters that pass all filtering and quality checks are used
to initialize new routing tokens and Procedural KV Memory blocks.
Noise points, unassigned samples, and rejected clusters are kept in
the episodic cache so that sparse tasks can continue accumulating
evidence over future streaming rounds.

\subsection{Text Representation}

For each boundary-agnostic sample, we construct a text representation that
captures both task-level and instance-level information. Let
$I_i$ denote the task instruction and $q_i$ denote the input query
of the $i$-th sample. We form the clustering text $x_i$ by
concatenating a truncated instruction prefix and a truncated query:
\begin{equation}
x_i =
\operatorname{Trunc}_{\lfloor L/2 \rfloor}(I_i)
\; || \;
\operatorname{Trunc}_{L-\lfloor L/2 \rfloor}(q_i),
\end{equation}
where $||$ denotes string concatenation and $L=500$ is the maximum
number of characters. This representation allocates approximately
half of the input budget to the instruction and half to the query,
ensuring that the clustering signal reflects both the abstract
task procedure and the concrete input pattern. All NCD computations
are performed over the resulting character-level text.

\subsection{NCD Distance Computation}

We use Normalized Compression Distance (NCD) to measure procedural
similarity between two samples. Given two clustering texts $x_i$
and $x_j$, NCD is defined as
\begin{equation}
\operatorname{NCD}(x_i, x_j)
=
\frac{
C(x_i x_j) - \min \{ C(x_i), C(x_j) \}
}{
\max \{ C(x_i), C(x_j) \}
},
\end{equation}
where $C(\cdot)$ denotes the compressed byte length computed by
\texttt{gzip}, and $x_i x_j$ denotes the concatenation of the two
texts. Smaller NCD values indicate higher similarity, while larger
values indicate lower similarity. In contrast to embedding-based
distances, NCD does not require a learned representation and is
sensitive to repeated lexical, syntactic, and structural patterns
that often characterize task-specific procedures.

We use two computation modes depending on the cache size
$n = |\mathcal{B}|$.

\paragraph{Full pairwise mode.}
When $n \leq 2000$, we compute the full pairwise NCD matrix
$D \in \mathbb{R}^{n \times n}$:
\begin{equation}
D_{ij} = \operatorname{NCD}(x_i, x_j).
\end{equation}
This mode preserves the exact pairwise distance structure among
all cached samples and is used as the default mode for most
clustering rounds. Its computational complexity is
$O(n^2)$ compression calls.

\paragraph{Anchor mode.}
When $n > 2000$, computing the full pairwise matrix becomes
expensive. We therefore switch to an anchor-based approximation.
Specifically, we randomly select $K=100$ anchor samples
$\{a_k\}_{k=1}^{K}$ from the current cache. Each sample $x_i$ is
represented by a $K$-dimensional NCD feature vector
$\mathbf{f}_i \in \mathbb{R}^{K}$:
\begin{equation}
f_{i,k} = \operatorname{NCD}(x_i, a_k).
\end{equation}
This reduces the number of compression calls from $O(n^2)$ to
$O(nK)$. The resulting feature matrix is then projected to 30
dimensions using UMAP with cosine distance and
\texttt{random\_state=42}. Clustering is performed in this reduced
feature space.

\subsection{HDBSCAN Clustering}

After constructing either the full NCD distance matrix or the
anchor-based feature representation, we apply HDBSCAN to discover
candidate latent tasks. HDBSCAN is suitable for the streaming
setting because it does not require specifying the number of
clusters in advance and can explicitly mark uncertain samples as
noise.

In full pairwise mode, HDBSCAN receives the precomputed NCD
distance matrix and uses \texttt{metric=precomputed}. In anchor
mode, HDBSCAN operates on the reduced feature representation and
uses Euclidean distance. Unless otherwise specified, we use the
following hyperparameters:
\begin{itemize}
    \item \texttt{min\_cluster\_size} $= S_{\min}$, with
    $S_{\min}=50$ by default;
    \item \texttt{min\_samples} $=100$, which makes cluster
    assignment conservative and reduces premature consolidation;
    \item \texttt{cluster\_selection\_method=eom}, corresponding
    to the Excess-of-Mass criterion.
\end{itemize}

Samples assigned the label $-1$ by HDBSCAN are treated as noise.
They are not used for parametric consolidation in the current
round and remain in the episodic cache for future clustering
rounds.

\subsection{Embedding-Based Outlier Filtering}

The raw clusters returned by HDBSCAN may still contain boundary
samples or semantically inconsistent samples. We therefore apply
an optional embedding-based outlier filtering step to each
candidate cluster. This step uses dense semantic representations
to complement the surface-form similarity captured by NCD.

For a candidate cluster with index set $\mathcal{I}$, we first
encode each sample using Qwen3-Embedding-0.6B and obtain
$\ell_2$-normalized embeddings
$\{\mathbf{e}_i\}_{i \in \mathcal{I}}$. The normalized cluster
centroid is computed as
\begin{equation}
\bar{\mathbf{e}}
=
\frac{
\sum_{i \in \mathcal{I}} \mathbf{e}_i
}{
\left\|
\sum_{i \in \mathcal{I}} \mathbf{e}_i
\right\|_2
}.
\end{equation}
For each sample, we compute its cosine similarity to the centroid:
\begin{equation}
s_i = \mathbf{e}_i^\top \bar{\mathbf{e}}.
\end{equation}
Let $\bar{s}$ and $\sigma_s$ denote the mean and standard deviation
of $\{s_i\}_{i \in \mathcal{I}}$. A sample is removed as an outlier
if
\begin{equation}
s_i < \bar{s} - k \cdot \sigma_s,
\end{equation}
where $k$ controls the sensitivity of the filter. We set
$k=0.5$ by default. After filtering, if the remaining cluster size
falls below $S_{\min}$, the entire cluster is discarded for the
current round and its samples are retained in the episodic cache.

\subsection{Cluster Quality Assessment}

After outlier filtering, we further evaluate the procedural
cohesion of each candidate cluster using intra-cluster NCD
similarity. For a cluster $\mathcal{C}$, let $\mathcal{P}$ denote
a set of unordered sample pairs drawn from $\mathcal{C}$. To reduce
computation, we use all pairs when $|\mathcal{C}| \leq 15$ and
otherwise randomly sample up to 50 pairs. The NCD-based cluster
similarity is defined as
\begin{equation}
\operatorname{sim}_{\mathrm{NCD}}(\mathcal{C})
=
1 -
\frac{1}{|\mathcal{P}|}
\sum_{(i,j) \in \mathcal{P}}
\operatorname{NCD}(x_i, x_j).
\end{equation}
A cluster is accepted only if
\begin{equation}
\operatorname{sim}_{\mathrm{NCD}}(\mathcal{C}) \geq \tau_{\mathrm{cluster}},
\end{equation}
where the default threshold is $\tau_{\mathrm{cluster}}=0.85$. This threshold is
controlled by the hyperparameter
\texttt{--cluster\_quality\_threshold}. Clusters with
$\operatorname{sim}_{\mathrm{NCD}}(\mathcal{C}) < \tau_{\mathrm{cluster}}$ are
considered low-quality and are excluded from supervised
fine-tuning in the current round. Their samples remain in the
episodic cache, allowing them to be reconsidered when additional
evidence becomes available.

\subsection{Streaming Cache Management}

At the end of each clustering round, UniMem updates the episodic
cache according to the clustering outcome. Samples belonging to
accepted clusters are removed from the cache and passed to the
procedural consolidation stage, where they are used to initialize
new routing tokens and train the corresponding Procedural KV
Memory blocks. Samples labeled as noise, samples removed by the
embedding-based outlier filter, and samples belonging to
low-quality clusters are retained in the cache.

This cache update rule is intentionally conservative. It prevents
UniMem from allocating parametric memory to one-off or ambiguous
patterns, while still allowing sparse tasks to accumulate enough
samples across multiple streaming rounds. As a result, the system
consolidates recurring and coherent task structures into
parametric memory, while continuing to serve uncertain or
long-tail samples through episodic retrieval.

\section{Computational Budget and Software Environment}
\label{app:compute_environment}

\paragraph{Computational budget.}
All experiments can be run on a single NVIDIA H20 GPU with 141GB memory. 
For the SNI streaming experiments, the computational cost depends mainly on the number of streaming tasks. 
A complete run with 100 SNI tasks takes approximately 3--4 hours on one H20 GPU. 
The 50-task setting takes approximately 2 hours, while the 10-task setting takes approximately 20 minutes. 
These estimates include streaming inference, episodic buffering, task discovery, and parametric consolidation. 
The reported runtime may vary slightly with backbone size, sequence length, and the number of discovered clusters.

\begin{table}[h]
\centering
\small
\begin{tabular}{lll}
\toprule
\textbf{Setting} & \textbf{Hardware} & \textbf{Approx. Runtime} \\
\midrule
SNI-10  & 1$\times$ NVIDIA H20 141GB & 20 minutes \\
SNI-50  & 1$\times$ NVIDIA H20 141GB & 2 hours \\
SNI-100 & 1$\times$ NVIDIA H20 141GB & 3--4 hours \\
\bottomrule
\end{tabular}
\caption{Approximate computational budget for the main SNI streaming experiments.}
\label{tab:compute_budget}
\end{table}

\paragraph{Software environment.}
Table~\ref{tab:software_environment} summarizes the main libraries used in our implementation. 
PyTorch and Transformers are used for model training, inference, tokenization, and generation. 
PEFT is used for the LoRA baseline. 
HDBSCAN and UMAP are used in the task discovery module. 
Sentence-Transformers is used for embedding-based retrieval and outlier filtering, and \texttt{rouge-score} is used for ROUGE-L F1 evaluation.
\begin{table}[h]
\centering
\small
\begin{tabular}{ll}
\toprule
\textbf{Library} & \textbf{Version} \\
\midrule
PyTorch & 2.8.0+cu128 \\
transformers & 4.57.1 \\
peft & 0.17.1 \\
hdbscan & 0.8.43 \\
umap-learn & 0.5.7 \\
scikit-learn & 1.7.2 \\
sentence-transformers & 5.1.1 \\
rouge-score & 0.1.2 \\
numpy & 1.26.4 \\
\bottomrule
\end{tabular}
\caption{Main software libraries used in our experiments.}
\label{tab:software_environment}
\end{table}

\section{Potential Risks}
\label{app:potential_risks}

Our experiments are conducted on public NLP benchmarks and do not collect, store, or process private user data. UniMem is studied as a research prototype for memory expansion in LLM agents rather than as a deployed user-facing system. Nevertheless, persistent memory mechanisms may introduce risks in real-world applications. Incorrect or biased strategies could be consolidated and repeatedly reused if the memory update process is not properly monitored. In addition, if such a system were applied to real user interactions, sensitive information could be unintentionally retained without appropriate filtering. These risks are not present in our benchmark experiments, but they should be addressed before deployment through privacy filtering, memory auditing, and conservative consolidation policies.

\section{Information About Use of AI Assistants}
\label{app:ai_assistants}

The authors used AI assistants for language polishing, writing refinement, and clarity checking during manuscript preparation. AI assistants were not used to generate experimental results or make final scientific claims. All technical content, experimental analyses, and final writing decisions were verified and controlled by the authors. AI assistants do not meet authorship criteria and are not listed as authors.

\end{document}